\documentclass[11pt,a4paper]{article}
\usepackage[hyperref]{acl2021}
\usepackage{times}
\usepackage{latexsym}
\usepackage{amsmath}
\usepackage{booktabs}
\usepackage{multirow}
\usepackage{graphicx}

\usepackage{microtype}

\aclfinalcopy 

\title{Similarity Based Label Smoothing For Dialogue Generation}

\author{Sougata Saha$^{*}$, Souvik Das\thanks{\ \ Equal Contribution}, Rohini Srihari\\
  Department of Computer Science and Engineering \\
  University at Buffalo, New York \\
  \texttt{\{sougatas, souvikda, rohini\}@buffalo.edu} \\}

\begin{document}
\maketitle
\begin{abstract}
Generative neural conversational systems are generally trained with the objective of minimizing the entropy loss between the training ``hard" targets and the predicted logits. Often, performance gains and improved generalization can be achieved by using regularization techniques like label smoothing, which converts the training ``hard" targets to ``soft" targets. However, label smoothing enforces a data independent uniform distribution on the incorrect training targets, which leads to an incorrect assumption of equi-probable incorrect targets for each correct target. In this paper we propose and experiment with incorporating data dependent word similarity based weighing methods to transforms the uniform distribution of the incorrect target probabilities in label smoothing, to a more natural distribution based on semantics. We introduce hyperparameters to control the incorrect target distribution, and report significant performance gains over networks trained using standard label smoothing based loss, on two standard open domain dialogue corpora.

\end{abstract}

\section{Introduction}
Generative conversational systems rely heavily on language modelling to be able to generate an appropriate response to a user query. Given a context which consist of multiple utterances in a conversation, a generative conversational system can be formulated as a next utterance prediction problem, where the task is to generate a response utterance conditioned on the context utterances. With the advent of deep learning and availability of sufficient training data, parametric models like recurrent neural networks and transformers are generally implemented to achieve the language modelling task. Trained by minimizing the expected cross entropy between the training hard targets and the prediction logits, such models often overfit the training data and does not generalize well on the test set. Label smoothing proposed by \cite{labelsmoothing} to improve the performance of Inception net image classifier on the ImageNet dataset, has widely gained acceptance in Natural Language Processing tasks as a regularization technique to enhance the generalization capability of deep neural networks. \cite{vaswani2017attention} in his work ``Attention is all you need", where he proposed the state-of-the-art transformer architecture, had reported performance gains in machine translation using label smoothing during training. Unlike other regularization techniques which constrain the model parameters and hidden representations, label smoothing augments the actual targets by reducing the actual target probability, and assigning low probabilities to all classes, following a data independent uniform distribution, thus preventing the model from predicting the correct labels too confidently during training. However, as pointed out by \cite{pereyra2017regularizing} and \cite{hinton2015distilling}, the probabilities assigned to both the correct and incorrect classes constitute the knowledge of a network. In language modelling, incorporating label smoothing and assigning a uniform probability to all the incorrect classes can convey an incorrect knowledge to the model. For example, in response to a user query ``How are you doing ?", if we want to generate the sentence ``I am doing good ." as the next utterance, given that we have already generated the phrase ``I am doing",  ``great" and ``awesome" can also convey the same message as ``good". On the other hand, ``bad" would convey a different message, but would be logically correct. However a random word like ``aeroplane" would not make any sense. Hence, if we want to use label smoothing, we would not want to follow a uniform distribution for the incorrect classes, and rather weigh them using a weighing mechanism which can present such knowledge to the model. In this paper we present ways of imparting such information by modifying the data independent uniform distribution in label smoothing with a more appropriate data dependent distribution, which is proportional to the pre-trained word embedding similarity between the actual target and the incorrect targets. 

\section{Related Work}
Numerous techniques have been introduced to enhance the generalization capability of neural networks. Although, as pointed out by \cite{pereyra2017regularizing}, substantial advancements have been made in regularizing model parameters, but not much work has been done in understanding external regularization techniques like label smoothing or target data augmentation. We can broadly categorize the most recent solutions to attain generalization in conversations in the following two categories.\\
\textbf{Loss function augmentation: }\cite{li-etal-2016-diversity} proposes using Maximum Mutual Information along with the Cross Entropy loss, in order to penalize bland responses like ``I do not know", which are frequent in conversational datasets. \cite{Jiang_2019} analysed that the Cross Entropy loss function prefers frequent tokens, which leads to generating bland responses. Hence they proposed augmenting the Cross Entropy Loss with a frequency based corpus dependent weighing mechanism, in order to yield diverse responses.  \cite{wang2020improving} experiments with using optimal transport to match sequences generated in the teacher and student modes, and increasing performance of student forced networks on the test dataset by reducing the gap between the two modes.\\
\textbf{Data augmentation: }\cite{cai2020data} demonstrated that conversational datasets generally don't exhibit coherence in query response pairs, which affect the Cross Entropy loss. They propose a training data augmentation module, which can not only replace words in the actual target response with similar words using BERT \cite{devlin-etal-2019-bert}, but also augment the style of the response, preserving the meaning. They further introduced a neural weighting mechanism, which can assign weights or importance to the augmented and golden training data, and report significant performance gains. \cite{kang2020improved} demonstrated that the log loss is not robust to noise, and hence proposed truncating the distribution of the training targets to achieve an easy to optimize and more robust loss function. \cite{he2020negative} introduced a network which can provide negative generated samples, and train the generation model to maximize the log likelihood of training data while minimizing the likelihood of negative samples. Our proposed method falls in the first category, as we do not augment the training data, and instead augment the probability of incorrect labels for each correct label.

\section{Methods and Experiments}
We experiment with ways to augment the data independent uniform distribution enforced by Label Smoothing. Let $U_i$ be an utterance consisting of words $\{w_j\}_{j=1}^{N}$, where $N$ is the number of words in the utterance. For each word $w_j$, in label smoothing a probability $1-s$ is assigned to the true label $w_j$, and a probability of $s$ (smoothing factor) is distributed uniformly among the rest of the $k$ words in the vocabulary. We augment the distribution of the incorrect class by weighting the smoothing factor $s$ according to the cosine similarity between the Glove \cite{pennington-etal-2014-glove} word embedding of the correct word in the training data and all the words in the vocabulary. Thus, if the correct word to be predicted is ``good", then the words ``great" and ``awesome" in the vocabulary would get a higher proportion of the smoothing factor $s$, compared to an unrelated word like ``aeroplane", thus presenting a more correct knowledge to the model. Mathematically, let $\vec{w_j}$ be the Glove word embedding of word $w_j$, $\mathbf{W_k}$ be a matrix containing the Glove word embedding for all the words in the vocabulary (including $w_j$),  $\vec{w}_{j\_sim}$ be the vector of cosine similarity between the word $w_j$ and all the words in the vocabulary. Since Glove word embeddings are learned representations, they can be noisy. Hence, we introduce a threshold $t$, below which we set the cosine similarity value in $\vec{w}_{j\_sim}$ as 0. We achieve this by introducing a mask $mask_j$, and multiply the similarity vector with the mask. The resulting vector is normalised to lie between 0 and 1, and finally multiplied by $s$. We treat $t$ as a model hyperparameter, and is tuned using grid search. We further reason that although Glove embeddings are learned from text corpora, there are possibilities that dissimilar words can lie in close proximity in the embedding space, resulting in a high cosine similarity score, and presenting an incorrect knowledge to the model. To circumvent this problem, we further experiment with filtering out the cosine similarities of dissimilar words based on WordNet sysnets \cite{princeton_university}, which we achieve by implementing another mask $mask_j^{'}$.
\begin{align*}
\vec{w}_{j\_sim} &= \frac{\vec{w_j} \cdot \mathbf{W_k}}{||\vec{w_j}||\cdot||\mathbf{W_k}||} \\
\vec{w}_{j\_dist} &= \frac{\vec{w}_{j\_sim} * mask_j * mask_j^{'}}{\sum (\vec{w}_{j\_sim} * mask_j * mask_j^{'})}*s \\
\vec{w}_{j\_dist}[j^*] &= 1-s \\
  \text{where, } mask_j &=\left\{
  \begin{array}{@{}ll@{}}
    0, & \text{if}\ \vec{w}_{j\_sim}<=t \\
    0, &\text{if}\ \vec{w}_{j\_sim}=1 \\
    1, & \text{otherwise}
  \end{array}\right.\\
  \text{and, } mask_j^{'} &=\left\{
  \begin{array}{@{}ll@{}}
    1, & \text{if $w_k$ is a synonym of $w_j$} \ \\
    0, & \text{otherwise}
  \end{array}\right.
\end{align*}

\subsection{Dataset}
We perform experiments on (i) The DailyDialog dataset \cite{li-etal-2017-dailydialog} : A multi-turn open domain dialogue dataset which has 13,118 pertaining to diverse day-to-day topics , and (ii) The Empathetic Dialogues dataset \cite{rashkin-etal-2019-towards} : An open domain multi-turn dataset consisting of 25,000 conversations grounded in emotional situations. We use the same training, validation and testing splits as mentioned in the datasets. We concatenate all the turns in the query in one long text, and use two special tokens: ``[speaker1]" and ``[speaker2]" to distinguish the speakers. In order to speed up computation, we restrict the context to the most recent 50 tokens, which is determined analytically from the corpora. Please refer to the supplementary material for the code and the dataset.

\subsection{Model}
Since the primary scope of this paper is to experiment with different loss functions, we used a standard transformer encoder-decoder architecture as proposed by \cite{vaswani2017attention}, where the encoder encodes the most recent utterance in the conversation, along with context from the previous turns. The encoder-decoder comprises of 3 layers each, with 300 dimensional hidden representation, with 6 attention heads in each multi-headed attention layer. The embedding layer is populated with 300 dimensional Glove  embeddings, which are trained along with the entire network. Finally, a fully connected linear layer predicts the next word.

\subsection{Experiments}
We treat the vanilla Cross Entropy (CE) loss, CE loss with label smoothing, Kullback–Leibler (KL) divergence loss and KL loss with label smoothing as the baselines. We experiment with different smoothing values $s \in \{NA, 0.1, 0.2\}$ , cosine similarity thresholds $t \in \{NA, 0.0, 0.5, 0.8\}$, and also perform ablation study to analyze the usefulness of the WordNet similarity mask $w \in \{NA, 0, 1\}$. In total, we experiment with 30 different settings for each dataset. 

\begin{figure}[h]
    \centering
    \includegraphics[width=7.7cm]{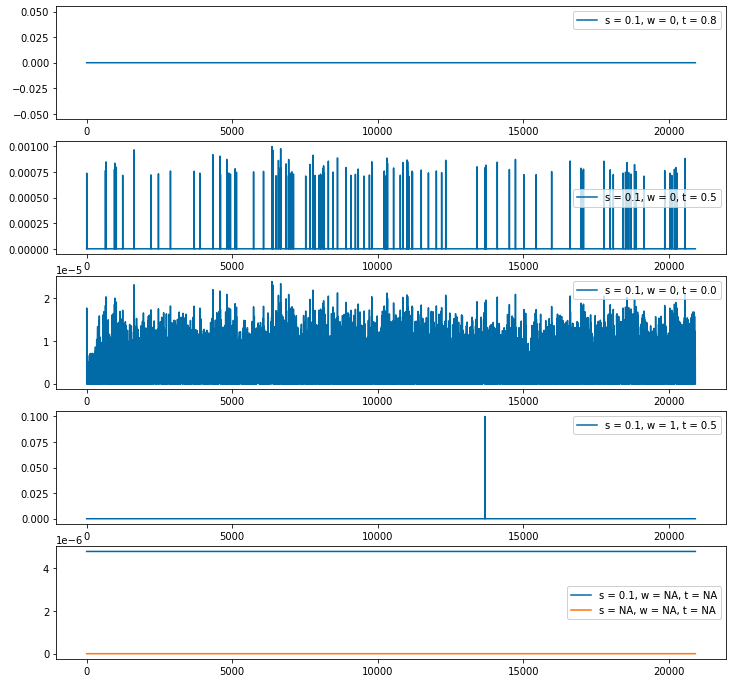}
    \caption{Illustration of probability of the incorrect words. The x-axis = vocabulary, y-axis = probability. Setting t = 0.8, or using WordNet mask filters out most words, making the target distribution equivalent to vanilla CE loss targets. Using t = 0.5 or 0.0 yields a less dramatic effect, and helps preserving the knowledge of the incorrect labels. }
    \label{fig:targets}
\end{figure}

\begin{table*}[!t]
\centering
\resizebox{\textwidth}{!}{%
\begin{tabular}{|c|c|c|ccc|cccccccc|}
\hline
\multicolumn{3}{|c|}{\multirow{2}{*}{}} & \textbf{s = NA} & \textbf{s = 0.1} & \textbf{s = 0.2} & \multicolumn{4}{c|}{\textbf{s =   0.1}} & \multicolumn{4}{c|}{\textbf{s = 0.2}} \\ \cline{4-14} 
\multicolumn{3}{|c|}{} & \textbf{t = NA} & \textbf{t = NA} & \textbf{t = NA} & \multicolumn{2}{c|}{\textbf{t = 0}} & \multicolumn{2}{c|}{\textbf{t = 0.5}} & \multicolumn{2}{c|}{\textbf{t = 0}} & \multicolumn{2}{c|}{\textbf{t = 0.5}} \\ \hline
\textbf{Dataset} & \textbf{Metric} & \textbf{Loss} & \textbf{w = NA} & \textbf{w = NA} & \textbf{w = NA} & \textbf{w = 0} & \textbf{w = 1} & \textbf{w = 0} & \textbf{w = 1} & \textbf{w = 0} & \textbf{w = 1} & \textbf{w = 0} & \textbf{w = 1} \\ \hline
\multirow{6}{*}{DD} & \multirow{2}{*}{SacreBLEU} & CE & 1.6625 & 1.8523 & 1.7251 & 1.9896 & 1.7627 & \textbf{\begin{tabular}[c]{@{}c@{}}2.1936 \\ (+ 12.67 \%)\end{tabular}} & 1.8575 & 2.1158 & 1.8020 & 2.0536 & 1.9302 \\
 &  & KL & \textbf{1.9469} & 1.7536 & 1.7931 & 1.8459 & 1.8181 & 1.9128 & 1.8858 & 1.9387 & 1.8092 & 1.7292 & 1.8856 \\ \cline{2-14} 
 & \multirow{2}{*}{ROUGE L} & CE & 0.1209 & 0.1243 & 0.1243 & 0.1238 & 0.1209 & \textbf{\begin{tabular}[c]{@{}c@{}}0.1270 \\ (+ 0.57 \%)\end{tabular}} & 0.1260 & 0.1217 & 0.1204 & 0.1238 & 0.1244 \\
 &  & KL & \textbf{0.1263} & 0.1223 & 0.1233 & 0.1227 & 0.1264 & 0.1243 & 0.1242 & 0.1223 & 0.1253 & 0.1232 & 0.1234 \\ \cline{2-14} 
 & \multirow{2}{*}{METEOR} & CE & 0.1244 & \textbf{0.1324} & 0.1286 & 0.1342 & 0.1287 & \textbf{\begin{tabular}[c]{@{}c@{}}0.1379 \\ (+ 4.16 \%)\end{tabular}} & 0.1314 & 0.1344 & 0.1279 & 0.1346 & 0.1313 \\
 &  & KL & \textbf{0.1324} & 0.1303 & 0.1303 & 0.1346 & 0.1324 & 0.1327 & 0.1311 & 0.1319 & 0.1310 & 0.1298 & 0.1296 \\ \hline
\multirow{6}{*}{ED} & \multirow{2}{*}{SacreBLEU} & CE & 2.2794 & \textbf{2.4084} & 2.1903 & \textbf{\begin{tabular}[c]{@{}c@{}}2.4427 \\ (+ 1.42 \%)\end{tabular}} & 2.2082 & 2.1922 & 2.2164 & 2.3187 & 2.2622 & 2.3125 & 2.2569 \\
 &  & KL & 2.2715 & 2.1682 & 2.2797 & 2.2774 & 2.2781 & 2.3370 & 2.3615 & 2.4319 & 2.2749 & 2.4393 & 2.1431 \\ \cline{2-14} 
 & \multirow{2}{*}{ROUGE L} & CE & 0.1382 & \textbf{0.1437} & 0.1373 & 0.1443 & 0.1409 & 0.1425 & 0.1385 & 0.1416 & 0.1398 & 0.1411 & 0.1381 \\
 &  & KL & 0.1406 & 0.1395 & 0.1426 & 0.1441 & 0.1454 & 0.1435 & 0.1387 & \textbf{\begin{tabular}[c]{@{}c@{}}0.1465 \\ (+ 1.95 \%)\end{tabular}} & 0.1401 & 0.1430 & 0.1394 \\ \cline{2-14} 
 & \multirow{2}{*}{METEOR} & CE & 0.1254 & 0.1287 & 0.1257 & \textbf{\begin{tabular}[c]{@{}c@{}}0.1324 \\ (+ 2.08 \%)\end{tabular}} & 0.1266 & 0.1248 & 0.1245 & 0.1291 & 0.1266 & 0.1278 & 0.1243 \\
 &  & KL & 0.1250 & 0.1233 & \textbf{0.1297} & 0.1272 & 0.1302 & 0.1290 & 0.1246 & 0.1323 & 0.1253 & 0.1283 & 0.1234 \\ \hline
\end{tabular}%
}
\caption{Comparison of sacreBLEU, ROUGE L and METEOR scores using variants of Cross Entropy (CE) loss and  Kullback–Leibler (KL) divergence loss on DailyDialog (DD) and EmpatheticDialogues (ED) datasets.}
\label{tab:results}
\end{table*}

\section{Results and Analysis}
We compare the (i) sacreBLEU score \cite{post-2018-call}: a standardised version of the BLEU score \cite{papineni-etal-2002-bleu}, (ii) ROUGE L score \cite{lin-2004-rouge}: which compares Longest Common Subsequence (LCS), and automatically takes into account sentence level structure similarity and identifies longest co-occurring in sequence n-grams, (iii) METEOR score \cite{banarjee2005}: an improvement over BLEU score, which incorporates stemming and synonymy matching along with exact word matching. Table \ref{tab:results} summarizes the results we obtained in each of the experiments. The supplementary material contains results for all the 60 experiments along with additional evaluation metrics like BERTscore \cite{bert-score} and ROUGE 1 \& 2. In Table \ref{tab:results}, the columns containing ``NA" are the baseline results, against which improvements are measured.\\
\textbf{Observations}
From the experiments we observe that, \textbf{(i)} Using a data dependent cosine similarity based distribution for label smoothing significantly outperforms the baseline (vanilla entropy based loss with or without label smoothing). We observe 12.67 \% increase in BLEU score, 0.57 \% increase in ROUGE L score, and 4.16 \% increase in METEOR score for the DailyDialog dataset, and 1.42 \% increase in BLEU score, 1.95 \% increase in ROUGE L score, and 2.08 \% increase in METEOR score for the EmpatheticDialogues dataset. \textbf{(ii)} Using additional WordNet synonym based filtering ($w$) does not help performance. To understand why this is happening, we plotted the distribution of the smoothing factor $s$ for the randomly selected word ``fun", and observed that the word had only one overlapping WordNet synonym in our vocabulary: ``play". This caused the word ``play" to be assigned a probability of 0.1, while all the other words are assigned a probability of 0, except for ``fun", which was assigned a probability of 0.9. We reason that the sparsity in synonyms does not help in reducing the overconfidence of the model, as the final distribution is very similar to non-smoothing targets. Figure \ref{fig:targets} illustrates the probabilities assigned to the incorrect labels of the word ``fun", by each of the methods discussed in this paper. \textbf{(iii)} Using CE loss instead of KL generally improves performance while using label smoothing. We reason that this happens because in case of label smoothing, the constant entropy coefficient in KL loss reduces the overall loss, thus reducing the gradients during back propagation, which results in slower learning. \textbf{(iv)} Generally, using high smoothing value ($s$) does not help in learning. \textbf{(v)} The cosine similarity threshold $t$ should be treated as a hyperparameter, and will require tuning depending on the vocabulary of the dataset used. \textbf{(vi)} We also noticed that a cosine similarity threshold $t$ as high as 0.8 does not help in learning. We reason that using a high threshold creates a scenario similar to using WordNet synonyms, where the smoothing probability is distributed among very few (or no) words. Note that in order to enhance readability, the results with 0.8 threshold are omitted from Table \ref{tab:results}, and are presented in the additional supplementary materials.

\section{Conclusion}
Label smoothing has an undesirable property of assigning uniform probability to incorrect labels, which present an incorrect knowledge to learn from. In this paper we propose ways to convert the uniform distribution to a data dependent distribution by weighing the smoothing probability using cosine similarity of word embeddings between the correct and incorrect labels. We further experiment with WordNet synonyms as an additional filtering criteria, and report our findings. Using our proposed methodology, we attain significant improvements over the baseline metrics across all datasets. However, one drawback that we notice in the proposed system is the inability to factor in context, while weighing the distribution of the incorrect labels. As future research, we intend to address this drawback using more contextualised representations instead of static embeddings. 

\bibliographystyle{acl_natbib}
\bibliography{acl2021}

\section{Supplementary Material}
\subsection{All Experiment Results}

Table \ref{tab:dd_ed_baselines} shows the different variants of the baselines that were computed for both the DailyDialog and EmpatheticDialogues datasets. All performance improvements are compared against these baselines. For a metric, the best baseline score among all the hyperparameter settings is chosen to report improvements. Table \ref{tab:dd_results} shows the results of using different hyperparameter settings and loss function in the DailyDialog dataset, and Table \ref{tab:ed_results} shows the results obtained on the EmpatheticDialogues dataset. The best results with detailed comparison against baselines are already discussed in the main paper.
\begin{table*}[b!]
\centering
\resizebox{10.5cm}{!}{%
\begin{tabular}{|cl|lll|lll|}
\hline
\multicolumn{1}{|l}{} &  & \multicolumn{3}{c|}{\textbf{DailyDialog Dataset}} & \multicolumn{3}{l|}{\textbf{EmpatheticDialogue Dataset}} \\ \cline{3-8} 
\multicolumn{1}{|l}{} &  & \textbf{s = NA} & \textbf{s = 0.1} & \textbf{s = 0.2} & \textbf{s = NA} & \textbf{s = 0.1} & \textbf{s = 0.2} \\ \cline{3-8} 
\multicolumn{1}{|l}{} &  & \textbf{t = NA} & \textbf{t = NA} & \textbf{t = NA} & \textbf{t = NA} & \textbf{t = NA} & \textbf{t = NA} \\ \hline
\multicolumn{1}{|l|}{\textbf{Metric}} & \textbf{Loss} & \textbf{w = NA} & \textbf{w = NA} & \textbf{w = NA} & \textbf{w = NA} & \textbf{w = NA} & \textbf{w = NA} \\ \hline
\multicolumn{1}{|c|}{\multirow{2}{*}{sacreBLEU}} & CE & 1.6625 & 1.8523 & 1.7251 & 2.2794 & 2.4084 & 2.1903 \\
\multicolumn{1}{|c|}{} & KL & 1.9469 & 1.7536 & 1.7931 & 2.2715 & 2.1682 & 2.2797 \\ \hline
\multicolumn{1}{|c|}{\multirow{2}{*}{BERTScore}} & CE & 0.8522 & 0.8520 & 0.8520 & 0.8539 & 0.8544 & 0.8527 \\
\multicolumn{1}{|c|}{} & KL & 0.8529 & 0.8520 & 0.8510 & 0.8540 & 0.8531 & 0.8541 \\ \hline
\multicolumn{1}{|c|}{\multirow{2}{*}{ROUGE 1}} & CE & 0.1272 & 0.1312 & 0.1319 & 0.1536 & 0.1592 & 0.1527 \\
\multicolumn{1}{|c|}{} & KL & 0.1336 & 0.1298 & 0.1300 & 0.1560 & 0.1545 & 0.1587 \\ \hline
\multicolumn{1}{|c|}{\multirow{2}{*}{ROUGE 2}} & CE & 0.0282 & 0.0303 & 0.0299 & 0.0251 & 0.0292 & 0.0251 \\
\multicolumn{1}{|c|}{} & KL & 0.0305 & 0.0283 & 0.0282 & 0.0267 & 0.0259 & 0.0271 \\ \hline
\multicolumn{1}{|c|}{\multirow{2}{*}{ROUGE L}} & CE & 0.1209 & 0.1243 & 0.1243 & 0.1382 & 0.1437 & 0.1373 \\
\multicolumn{1}{|c|}{} & KL & 0.1263 & 0.1223 & 0.1233 & 0.1406 & 0.1395 & 0.1426 \\ \hline
\multicolumn{1}{|c|}{\multirow{2}{*}{METEOR}} & CE & 0.1244 & 0.1324 & 0.1286 & 0.1254 & 0.1287 & 0.1257 \\
\multicolumn{1}{|c|}{} & KL & 0.1324 & 0.1303 & 0.1303 & 0.1250 & 0.1233 & 0.1297 \\ \hline
\end{tabular}%
}
\caption{Baseline results of diverse automatic text generation metrics on the DailyDialog and EmpatheticDialogues datasets. The hyperparameters s, t and w control the usage of Label Smoothing, Cosine similarity threshold and WordNet filtering respectively. For the baseline, t and w were not used, which is indicated by NA. s = NA signifies vanilla entropy based loss without Label Smoothing.}
\label{tab:dd_ed_baselines}
\end{table*}

\subsection{Model Training and Parameters}
All the models were trained on a single Nvidia V-100 GPU, for 15 epochs each with a learning rate of 2e-4, batch size of 64, and using AdamW optimizer. The gradients of the model were clipped with a value of 1, and dropout with probability 0.1 was applied during training. The average run-time of each experiment is 60 minutes, with each of the trained models having 17.7 M parameters. The code, dataset and best performing models are publicly available through this link: \href{https://drive.google.com/file/d/1--w7x3QY1rbx4Byfb9DW4pUlVEK1s277/view?usp=sharing}{download link}.

\begin{table*}[t!]
\centering
\resizebox{\textwidth}{!}{%
\begin{tabular}{|cl|llllllllllll|}
\hline
\multicolumn{1}{|l}{\textbf{}} & \textbf{} & \multicolumn{6}{c|}{\textbf{s = 0.1}} & \multicolumn{6}{c|}{\textbf{s = 0.2}} \\ \cline{3-14} 
\multicolumn{1}{|l}{\textbf{}} & \textbf{} & \multicolumn{2}{c|}{\textbf{t = 0}} & \multicolumn{2}{c|}{\textbf{t = 0.5}} & \multicolumn{2}{c|}{\textbf{t = 0.8}} & \multicolumn{2}{c|}{\textbf{t = 0}} & \multicolumn{2}{c|}{\textbf{t = 0.5}} & \multicolumn{2}{c|}{\textbf{t = 0.8}} \\ \hline
\multicolumn{1}{|l|}{\textbf{Metric}} & \textbf{Loss} & \multicolumn{1}{c}{\textbf{w = 0}} & \multicolumn{1}{c}{\textbf{w = 1}} & \multicolumn{1}{c}{\textbf{w = 0}} & \multicolumn{1}{c}{\textbf{w = 1}} & \multicolumn{1}{c}{\textbf{w = 0}} & \multicolumn{1}{c}{\textbf{w = 1}} & \multicolumn{1}{c}{\textbf{w = 0}} & \multicolumn{1}{c}{\textbf{w = 1}} & \multicolumn{1}{c}{\textbf{w = 0}} & \multicolumn{1}{c}{\textbf{w = 1}} & \multicolumn{1}{c}{\textbf{w = 0}} & \multicolumn{1}{c|}{\textbf{w = 1}} \\ \hline
\multicolumn{1}{|c|}{\multirow{2}{*}{sacreBLEU}} & CE & 1.9896 & 1.7627 & 2.1936 & 1.8575 & 1.6676 & 1.8859 & 2.1158 & 1.8020 & 2.0536 & 1.9302 & 1.5674 & 1.8502 \\
\multicolumn{1}{|c|}{} & KL & 1.8459 & 1.8181 & 1.9128 & 1.8858 & 1.7957 & 1.7453 & 1.9387 & 1.8092 & 1.7292 & 1.8856 & 1.5874 & 1.9707 \\ \hline
\multicolumn{1}{|c|}{\multirow{2}{*}{BERTScore}} & CE & 0.8518 & 0.8529 & 0.8515 & 0.8527 & 0.8507 & 0.8509 & 0.8513 & 0.8525 & 0.8525 & 0.8519 & 0.8507 & 0.8512 \\
\multicolumn{1}{|c|}{} & KL & 0.8520 & 0.8527 & 0.8517 & 0.8515 & 0.8520 & 0.8516 & 0.8509 & 0.8525 & 0.8522 & 0.8518 & 0.8518 & 0.8515 \\ \hline
\multicolumn{1}{|c|}{\multirow{2}{*}{ROUGE 1}} & CE & 0.1309 & 0.1279 & 0.1353 & 0.1326 & 0.1260 & 0.1280 & 0.1298 & 0.1271 & 0.1315 & 0.1317 & 0.1250 & 0.1290 \\
\multicolumn{1}{|c|}{} & KL & 0.1301 & 0.1332 & 0.1318 & 0.1311 & 0.1281 & 0.1276 & 0.1301 & 0.1328 & 0.1310 & 0.1312 & 0.1263 & 0.1325 \\ \hline
\multicolumn{1}{|c|}{\multirow{2}{*}{ROUGE 2}} & CE & 0.0282 & 0.0276 & 0.0309 & 0.0300 & 0.0287 & 0.0280 & 0.0286 & 0.0286 & 0.0308 & 0.0310 & 0.0276 & 0.0305 \\
\multicolumn{1}{|c|}{} & KL & 0.0283 & 0.0312 & 0.0300 & 0.0294 & 0.0297 & 0.0291 & 0.0285 & 0.0312 & 0.0292 & 0.0299 & 0.0277 & 0.0299 \\ \hline
\multicolumn{1}{|c|}{\multirow{2}{*}{ROUGE L}} & CE & 0.1238 & 0.1209 & 0.1270 & 0.1260 & 0.1200 & 0.1203 & 0.1217 & 0.1204 & 0.1238 & 0.1244 & 0.1183 & 0.1222 \\
\multicolumn{1}{|c|}{} & KL & 0.1227 & 0.1264 & 0.1243 & 0.1242 & 0.1213 & 0.1207 & 0.1223 & 0.1253 & 0.1232 & 0.1234 & 0.1185 & 0.1252 \\ \hline
\multicolumn{1}{|c|}{\multirow{2}{*}{METEOR}} & CE & 0.1342 & 0.1287 & 0.1379 & 0.1314 & 0.1270 & 0.1319 & 0.1344 & 0.1279 & 0.1346 & 0.1313 & 0.1223 & 0.1280 \\
\multicolumn{1}{|c|}{} & KL & 0.1346 & 0.1324 & 0.1327 & 0.1311 & 0.1262 & 0.1275 & 0.1319 & 0.1310 & 0.1298 & 0.1296 & 0.1247 & 0.1330 \\ \hline
\end{tabular}%
}
\caption{Results of diverse automatic text generation metrics on the DailyDialog dataset, trained with variants of Entropy based loss with different hyperparameter settings: cosine similarity threshold (t), Label Smoothing (s) and WordNet filtering (w).}
\label{tab:dd_results}
\end{table*}

\begin{table*}[t!]
\centering
\resizebox{\textwidth}{!}{%
\begin{tabular}{|cl|llllllllllll|}
\hline
\multicolumn{1}{|l}{\textbf{}} & \textbf{} & \multicolumn{6}{c|}{\textbf{s = 0.1}} & \multicolumn{6}{c|}{\textbf{s = 0.2}} \\ \cline{3-14} 
\multicolumn{1}{|l}{\textbf{}} & \textbf{} & \multicolumn{2}{c|}{\textbf{t = 0}} & \multicolumn{2}{c|}{\textbf{t = 0.5}} & \multicolumn{2}{c|}{\textbf{t = 0.8}} & \multicolumn{2}{c|}{\textbf{t = 0}} & \multicolumn{2}{c|}{\textbf{t = 0.5}} & \multicolumn{2}{c|}{\textbf{t = 0.8}} \\ \hline
\multicolumn{1}{|l|}{\textbf{Metric}} & \textbf{Loss} & \multicolumn{1}{c}{\textbf{w = 0}} & \multicolumn{1}{c}{\textbf{w = 1}} & \multicolumn{1}{c}{\textbf{w = 0}} & \multicolumn{1}{c}{\textbf{w = 1}} & \multicolumn{1}{c}{\textbf{w = 0}} & \multicolumn{1}{c}{\textbf{w = 1}} & \multicolumn{1}{c}{\textbf{w = 0}} & \multicolumn{1}{c}{\textbf{w = 1}} & \multicolumn{1}{c}{\textbf{w = 0}} & \multicolumn{1}{c}{\textbf{w = 1}} & \multicolumn{1}{c}{\textbf{w = 0}} & \multicolumn{1}{c|}{\textbf{w = 1}} \\ \hline
\multicolumn{1}{|c|}{\multirow{2}{*}{sacreBLEU}} & CE & 2.4427 & 2.2082 & 2.1922 & 2.2164 & 2.3467 & 2.2596 & 2.3187 & 2.2622 & 2.3125 & 2.2569 & 2.3944 & 2.2767 \\
\multicolumn{1}{|c|}{} & KL & 2.2774 & 2.2781 & 2.3370 & 2.3615 & 2.2347 & 2.2769 & 2.4319 & 2.2749 & 2.4393 & 2.1431 & 2.2566 & 2.2652 \\ \hline
\multicolumn{1}{|c|}{\multirow{2}{*}{BERTScore}} & CE & 0.8543 & 0.8539 & 0.8547 & 0.8528 & 0.8536 & 0.8544 & 0.8544 & 0.8536 & 0.8539 & 0.8532 & 0.8531 & 0.8544 \\
\multicolumn{1}{|c|}{} & KL & 0.8541 & 0.8543 & 0.8544 & 0.8528 & 0.8536 & 0.8528 & 0.8544 & 0.8528 & 0.8544 & 0.8526 & 0.8535 & 0.8543 \\ \hline
\multicolumn{1}{|c|}{\multirow{2}{*}{ROUGE 1}} & CE & 0.1612 & 0.1564 & 0.1577 & 0.1531 & 0.1558 & 0.1551 & 0.1589 & 0.1550 & 0.1575 & 0.1531 & 0.1553 & 0.1590 \\
\multicolumn{1}{|c|}{} & KL & 0.1594 & 0.1613 & 0.1596 & 0.1540 & 0.1549 & 0.1552 & 0.1619 & 0.1554 & 0.1588 & 0.1545 & 0.1564 & 0.1569 \\ \hline
\multicolumn{1}{|c|}{\multirow{2}{*}{ROUGE 2}} & CE & 0.0287 & 0.0270 & 0.0271 & 0.0250 & 0.0274 & 0.0267 & 0.0269 & 0.0265 & 0.0267 & 0.0264 & 0.0261 & 0.0262 \\
\multicolumn{1}{|c|}{} & KL & 0.0270 & 0.0290 & 0.0273 & 0.0266 & 0.0256 & 0.0253 & 0.0288 & 0.0269 & 0.0274 & 0.0257 & 0.0251 & 0.0268 \\ \hline
\multicolumn{1}{|c|}{\multirow{2}{*}{ROUGE L}} & CE & 0.1443 & 0.1409 & 0.1425 & 0.1385 & 0.1402 & 0.1388 & 0.1416 & 0.1398 & 0.1411 & 0.1381 & 0.1396 & 0.1423 \\
\multicolumn{1}{|c|}{} & KL & 0.1441 & 0.1454 & 0.1435 & 0.1387 & 0.1397 & 0.1393 & 0.1465 & 0.1401 & 0.1430 & 0.1394 & 0.1404 & 0.1416 \\ \hline
\multicolumn{1}{|c|}{\multirow{2}{*}{METEOR}} & CE & 0.1324 & 0.1266 & 0.1248 & 0.1245 & 0.1267 & 0.1264 & 0.1291 & 0.1266 & 0.1278 & 0.1243 & 0.1247 & 0.1292 \\
\multicolumn{1}{|c|}{} & KL & 0.1272 & 0.1302 & 0.1290 & 0.1246 & 0.1235 & 0.1254 & 0.1323 & 0.1253 & 0.1283 & 0.1234 & 0.1257 & 0.1269 \\ \hline
\end{tabular}%
}
\caption{Results of diverse automatic text generation metrics on the EmpatheticDialogues dataset, trained with variants of Entropy based loss with different hyperparameter settings: cosine similarity threshold (t), Label Smoothing (s) and WordNet filtering (w).}
\label{tab:ed_results}
\end{table*}

\end{document}